%% file: olivetti_streamline_distance.tex
\documentclass[10pt, conference, compsocconf]{IEEEtran}
\IEEEoverridecommandlockouts

\usepackage{cite}

\usepackage{subfigure}
\ifCLASSINFOpdf
  \usepackage[pdftex]{graphicx}
\else
\fi

\usepackage[cmex10]{amsmath}
\usepackage{amsfonts}   

\usepackage[noend]{algorithmic}

\usepackage{array}
\usepackage{multirow}
\usepackage{url}

\newcommand{\argmin}{\operatornamewithlimits{argmin}}

\newcommand{\NN}{\operatorname{{\mathrm NN}}}

\begin{document}
%
\title{Comparison of Distances for Supervised Segmentation of White
  Matter Tractography}



 \author{\IEEEauthorblockN{Emanuele
     Olivetti\IEEEauthorrefmark{1}\IEEEauthorrefmark{2},
     Giulia Bert\`o\IEEEauthorrefmark{1}\IEEEauthorrefmark{2},
     Pietro Gori\IEEEauthorrefmark{3}, 
     Nusrat Sharmin\IEEEauthorrefmark{1}\IEEEauthorrefmark{2} and
     Paolo Avesani\IEEEauthorrefmark{1}\IEEEauthorrefmark{2}
   }
   \IEEEauthorblockA{\IEEEauthorrefmark{1}NeuroInformatics Laboratory (NILab),
     Bruno Kessler Foundation, Trento, Italy}
   \IEEEauthorblockA{\IEEEauthorrefmark{2}Centro Interdipartimentale
     Mente e Cervello (CIMeC),
     University of Trento, Italy}
   \IEEEauthorblockA{\IEEEauthorrefmark{3}Image processing and
     understanding (TII) group, T\'el\'ecom ParisTech, France}
   \thanks{The research was funded by the Autonomous Province of Trento,
     Call "Grandi Progetti 2012", project "Characterizing and improving
     brain mechanisms of attention - ATTEND".  }
 }



%


\maketitle

\input{abstract}


%
\IEEEpeerreviewmaketitle

\input{introduction}

\input{methods}

\input{experiments}

\input{discussion}




\bibliographystyle{IEEEtran}
\bibliography{group-prni2017distances-7255}

\end{document}

%% file: abstract.tex
\begin{abstract}
  Tractograms are mathematical representations of the main paths of
  axons within the white matter of the brain, from diffusion MRI
  data. Such representations are in the form of polylines, called
  streamlines, and one streamline approximates the common path of tens
  of thousands of axons. The analysis of tractograms is a task of
  interest in multiple fields, like neurosurgery and neurology. A
  basic building block of many pipelines of analysis is the definition
  of a distance function between streamlines. Multiple distance
  functions have been proposed in the literature, and different
  authors use different distances, usually without a specific reason
  other than invoking the ``common practice''. To this end, in this
  work we want to test such common practices, in order to obtain
  factual reasons for choosing one distance over another. For these
  reason, in this work we compare many streamline distance functions
  available in the literature. We focus on the common task of
  automatic bundle segmentation and we adopt the recent approach of
  supervised segmentation from expert-based examples. Using the HCP
  dataset, we compare several distances obtaining guidelines on the
  choice of which distance function one should use for supervised
  bundle segmentation.
\end{abstract}


%% file: introduction.tex
\section{Introduction}
\label{sec:introduction}

Current diffusion magnetic resonance imaging (dMRI) techniques,
together with tractography algorithms, allow the in-vivo
reconstruction of the main white matter pathways of the brain at the
millimiter scale, see~\cite{catani2015atlas}. The most common
representation of the white matter is in terms of 3D polylines, called
\emph{streamlines}, where one streamline approximates the path of tens
of thousands of axons sharing a similar path. The whole set of
streamlines of a brain is called \emph{tractogram} and it is usually
composed of $10^5-10^6$ streamlines.

In multiple applications, like neurosurgical planning and the study of
neurological disorders, tractograms are manipulated by algorithms 
to support navigation, quantification and virtual dissection,
performed by experts, see~\cite{catani2002virtual}. During the virtual
dissection of a tractogram, a given anatomical bundle of interest is
segmented by identifying the streamlines that approximate it
best. Such segmentation can be manual, e.g. by manually defining
regions of interest (ROIs) crossed by those streamlines, or fully
automated, like in the case of unsupervi-\\sed clustering~\cite{garyfallidis2012quickbundles} or supervised
segmentation~\cite{yoo2015example,sharmin2016alignment}.

A common basic building block for such algorithms is the definition of
a streamline-streamline distance function, to quantify the relative
displacements between streamlines. The idea is that streamlines
belonging to the same anatomical structure lie at small distances,
while streamlines belonging to different anatomical structures lie at
greater distances. The specific distance function defines the result
of nearest neighbor algorithm applied to a streamline. Such algorithm
is used in supervised bundle segmentation, see~\cite{yoo2015example},
where an example bundle of a subject is provided in order to learn how
to segment the same bundle in the tractogram of another subject.

In the literature, several streamline-streamline distance functions
have been proposed. The most common distances rely on streamlines
parametrized as sequences of 3D points, even though other
parametrizations exist such as B-splines~\cite{corouge2006fiber} or
Fourier descriptors~\cite{batchelor2006quantification}. This kind of
distances can then be separated into two main groups: those based on a
point-to-point correspondence between streamlines,
i.e. minimum-average direct flip (MDF)~\cite{garyfallidis2015robust},
and those not requiring that (i.e. Hausdorff,
currents~\cite{gori2016parsimonious}).

Even though each group of distances has a distinct technical
motivation, little has been said to guide the choice of the
practitioner when choosing a distance for a specific task. To the best
of our knowledge, only in the case of unsupervised bundle
segmentation, by means of clustering, some results are available about
comparing distances. In~\cite{moberts2005evaluation}, four different
distances were compared to see the impact on various indexes for
clustering of streamlines. In~\cite{siless2013comparison}, for the
task of clustering of streamlines, three distances have been compared,
obtaining some evidence that the point density model (PDM) distance
should be preferred for that task.

In this work, we propose to address the gap in the literature by
providing guidelines for the choice of the streamline-streamline
distance function for the specific task of supervised bundle
segmentation. Following the ideas
in~\cite{yoo2015example,sharmin2016alignment,olivetti2016alignment},
we adopt the supervised segmentation framework, where the desired
bundle is automatically segmented from a tractogram starting from an
example of that bundle segmented by an expert on a different subject.


We computed the supervised segmentations of 9 bundles with the nearest
neighbor algorithm using 8 different distance functions. We compared
the obtained bundles first against a ground truth, and then one against each other.

Our results show that the quality of segmented bundles does not
significantly change when changing the distance function, despite the
large differences in computational cost. At the same time, we observe
that, at the streamline level, different distances result in a
different nearest neighbor.

In the following, we first briefly introduce the notation, the
streamline distances and the the approximate nearest neighbor
algorithms used in this work. In Section~\ref{sec:experiments}, we
describe the details of the experimental setup and provide the
results. In Section~\ref{sec:discussion}, we discuss the results and
draw the conclusions.


%% file: methods.tex
\section{Methods}
\label{sec:methods}
Let $s = \{\mathbf{x}_1, \dots, \mathbf{x}_n\}$ be a streamline,
i.e. a sequence of points, where
$\mathbf{x}_i = [x_i, y_i, z_i] \in \mathbb{R}^3$, $\forall i$. Let
$T = \{s_1, \dots, s_N\}$ be a tractogram and let $b \subset T$
represent the set of streamlines corresponding to an anatomical bundle
of interest, e.g. the arcuate fasciculus. Usually, $n$ differs from
streamline to streamline, assuming values in the order of
$10^1-10^2$. We indicate the number of points of a streamline $s$ with
$|s|$. $N$ is usually in the order to $10^5-10^6$, depending on the
parameters of acquisistion of dMRI data and on reconstruction/tracking
algorithms.

\subsection{Streamline distances}
\label{sec:distances}
Here we define the streamline distances that are frequently used in
the literature and that are compared in this work.
\begin{itemize}
\item Mean of closest distances~\cite{zhang2008identifying}:
  \begin{equation}
    \label{eq:mam_avg}
    d_{MC}(s_a, s_b) = \frac{d_m(s_a, s_b) + d_m(s_b, s_a)}{2} 
  \end{equation}
  where
    $d_m(s_a, s_b) = \frac{1}{|s_a|} \sum_{\mathbf{x}_i \in s_a}
    \min_{\mathbf{x}_j \in s_b} ||\mathbf{x}_i - \mathbf{x}_j||_2$
\item Shorter mean of closest distances~\cite{zhang2008identifying}:
  \begin{equation}
    \label{eq:mam_min}
    d_{\text{SC}}(s_a, s_b) = \min(d_m(s_a, s_b), d_m(s_b, s_a))
  \end{equation}
\item Longer mean of closest distances~\cite{zhang2008identifying}:
  \begin{equation}
    \label{eq:mam_max}
    d_{LC}(s_a, s_b) = \max(d_m(s_a, s_b), d_m(s_b, s_a))
  \end{equation}
\item After re-sampling each streamline to a given number of points
  $m$, such as $s_a = \{\mathbf{x}_1^a, \dots, \mathbf{x}_m^a\}$ and
  $s_b = \{\mathbf{x}_1^b, \dots, \mathbf{x}_m^b\}$, the MDF distance,
  see~\cite{garyfallidis2015robust} is defined as:
  \begin{equation}
    \label{eq:mdf}
    d_{\text{MDF},m}(s_a, s_b) = \min (d_{\text{direct}}(s_a, s_b),
    d_{\text{flipped}}(s_a, s_b))
  \end{equation}
where
    $d_{\text{direct}}(s_a, s_b) = \frac{1}{m}\sum_{i=1}^m
    ||\mathbf{x}_i^a - \mathbf{x}_i^b ||_2$
and
    $d_{\text{flipped}}(s_a, s_b) = \frac{1}{m}\sum_{i=1}^m
    ||\mathbf{x}_i^a - \mathbf{x}_{m-i+1}^b ||_2$
\item Point Density Model (PDM, see~\cite{siless2013comparison}):
  \begin{equation}
    \label{eq:pdm}
    d_{\text{PDM}}^2(s_a, s_b) =  \langle s_a, s_a \rangle_{pdm} +
    \langle s_b, s_b \rangle_{pdm} -2 \langle s_a, s_b \rangle_{pdm}
  \end{equation}
where
\begin{equation}
  \label{eq:pdm_kernel}
  \langle s_a, s_b \rangle_{pdm} = \frac{1}{|s_a||s_b|}
  \sum_{i=1}^{|s_a|} \sum_{j=1}^{|s_b|} K_{\sigma}(\mathbf{x}_i^a,
  \mathbf{x}_j^b)
\end{equation}
and
$K_{\sigma}(\mathbf{x}_i^a, \mathbf{x}_j^b) = \exp
\left(-\frac{||\mathbf{x}_i^a -
    \mathbf{x}_j^b||_2^2}{\sigma^2}\right)$
is a Gaussian kernel between the two 3D points.

\item Varifolds distance (see~\cite{charon2013varifold}) is the
  non-oriented version of the currents\cite{gori2016parsimonious}
  distance, namely it does not need streamlines $a$ and $b$ to have a
  consistent orientation.
\begin{equation}
    \label{eq:varifolds}
 d_{\text{varifolds}}^2(s_a, s_b) =  \langle s_a, s_a \rangle_{var} + \langle s_b, s_b \rangle_{var} -2 \langle s_a, s_b \rangle_{var}
   \end{equation}
where
  \begin{equation}
    \label{eq:varifolds_inner}
    \langle s_a, s_b \rangle_{var} = \sum_{i=1}^{|s_a|-1}
    \sum_{j=1}^{|s_b|-1} K_{\sigma}(\mathbf{p}_i^a , \mathbf{p}_j^b)
    K_n(\mathbf{n}_i^a , \mathbf{n}_j^b)  | \mathbf{n}_i^a |_2 |
    \mathbf{n}_j^b |_2
  \end{equation}
  with $K_n(\mathbf{n}_i^a , \mathbf{n}_j^b)$ =
  $\left( \frac{(\mathbf{n}_i^a)^T \mathbf{n}_j^b}{| \mathbf{n}_i^a
      |_2 | \mathbf{n}_j^b |_2} \right)^2$
  where $\mathbf{p}_i^a$ (resp. $\mathbf{p}_j^b$) and $\mathbf{n}_i^a$
  (resp. $\mathbf{n}_j^b$) are the center and tangent vector of
  segment $i$ (resp. $j$) of streamline $a$ (resp. $b$). The endpoints
  of segment $i$ are $\mathbf{x}_i$ and $\mathbf{x}_{i+1}$ for
  $i \in [1,...,n-1]$.
\end{itemize}

\subsection{Supervised Segmentation of Bundles}
As in~\cite{yoo2015example,sharmin2016alignment}, we segment a bundle
of interest in the tractogram of a given (target) subject using a
supervised procedure. This means that we leverage the segmentation of
the same bundle in the tractogram of another subject, as an
example. Assuming that the tractograms of the two subjects are
registered in the same space, e.g. see~\cite{garyfallidis2015robust},
a simple supervised segmentation method is based on the nearest
neighbor algorithm: we define the segmented bundle as the set of
streamlines of the target subject that are nearest neighbor of the
streamlines of the example bundle.

More formally, let $T_{\text{example}}^A$ and $T_{\text{target}}^B$ be
the tractograms of two different subjects, $A$ and $B$. Let
$b_{\text{example}}^A \subset T_{\text{example}}^A$ be an example of
the bundle of interest, segmented by an expert. Let
$b_{\text{target}}^B \subset T_{\text{target}}^B$ be the (unknown)
corresponding bundle we want to approximate using automatic supervised
segmentation, via nearest neighbor. Under the assumptions that
$T_{\text{example}}^A$ and $T_{\text{target}}^B$ are co-registered,
the approximate bundle
$\hat{b}_{\text{target}}^B \subset T_{\text{target}}^B$ is such that
\begin{equation}
\label{eq:nn_segmentation}
\hat{b}_{\text{target}}^B = \{\NN(s_e^A, T_{\text{target}}^B), \forall s_e^A
\in b_{\text{example}}^A \}
\end{equation}
where $s_e^A$ is a streamline of the example tract of subject $A$ and
$\NN(s_e^A, T_{\text{target}}^B) = \argmin_{s^B \in
  T_{\text{target}}^B} d(s_e^A, s^B)$
its nearest neighbor streamline in $T_{\text{target}}^B$, i.e. the one
having minimum distance from $s_e^A$.

The notion of streamline-streamline distance can be implemented in
multiple ways, such as those listed above, in
Section~\ref{sec:distances}. For this reason, different distances
induce different segmentations.

Notice that, in principle, computing the nearest neighbors of the
streamlines in $T_{\text{target}}^B$ is expensive, in terms of
computations. The most basic algorithm would require the computation of
$|b_{\text{example}}^A| \times |T_{\text{target}}^B|$ distances, which
is usually in the order of $10^7-10^9$. According to the timings in
Table~\ref{tab:times}, a single nearest neighbors segmentation may
require over 24 hours of computation, in case of a large bundle.

\subsection{Efficient Computation of Nearest Neighbor}
\label{sec:efficient_nn}
Based on the results in~\cite{olivetti2012approximation}, we adopt a
simple procedure to efficiently compute the approximate nearest
neighbor of a streamline, that reduces the amount of computations of
several orders of magnitudes with respect to the standard
algorithm. The procedure is the following: first, we transform each
streamline in $T_{\text{target}}$ into an $d$-dimensional vector,
using an Euclidean embedding technique called \emph{dissimilarity
  representation}~\cite{pekalska2005dissimilarity}. For lack of space,
we refer the reader to~\cite{olivetti2012approximation} for all the
details. Second, we put all vectors in a $k$-d
tree~\cite{bentley1975multidimensional}, which is a space partitioning
data structure that provides efficient $1$-nearest neighbor search,
which requires only $\mathcal{O}(\log N)$ steps,
$N = |T_{\text{target}}|$. Then, for each streamline in
$b_{\text{example}}$, we transform it into a vector using again the
dissimilarity representation step above and we compute its nearest
neighbor in $T_{\text{target}}$ through the $k$-d tree.


%% file: experiments.tex
\section{Experiments}
\label{sec:experiments}

We conducted multiple experiments on the the Human Connectome Project
(HCP) dMRI datasets,
see~\cite{vanessen2013wuminn,sotiropoulos2013effects}, ($90$
gradients; b = $1000$; voxel size = ($1.25$ x $1.25$ x $1.25$
$mm^3$)). The reconstruction step was performed using the constrained
spherical deconvolution (CSD) algorithm~\cite{tournier2007robust} and
the tracking step using the Euler Delta Crossing (EuDX)
algorithm~\cite{garyfallidis2012quickbundles} with $10^6$ seeds. We
adopted the white matter query language
(WMQL)~\cite{wassermann2013describing} to obtain 9 segmented bundles
for 10 random subjects, which we considered as ground truth. We
selected the bundles reproducing the selection
in~\cite{olivetti2016alignment}, where they aimed to avoid extreme
variability of the same bundle across subjects, due to the limitations
of WMQL. The selected bundles are reported in the first column of Table~\ref{tab:main_results}.
Each pair of tractograms was co-registered
using the streamline linear registration (SLR)
algorithm~\cite{garyfallidis2015robust}.

As explained in Section~\ref{sec:efficient_nn}, we represented the
streamlines into a vectorial space, in order to obtain fast nearest
neighbor queries. We considered 8 different distance functions,
described in Section~\ref{sec:methods}: $d_{\text{MC}}$,
$d_{\text{SC}}$, $d_{\text{LC}}$, $d_{\text{MDF},12}$,
$d_{\text{MDF},20}$, $d_{\text{MDF},32}$, $d_{\text{PDM}}$ and
$d_{\text{varifolds}}$. For $d_{\text{PDM}}$ and
$d_{\text{varifolds}}$ we set $\sigma=42mm$, according
to~\cite{siless2013comparison}. For each subject and distance
function, we computed the dissimilarity representation of the (target)
tractogram $T_{\text{target}}^B$. According
to~\cite{olivetti2012approximation}, we selected 40 prototypes with
the subset farthest first (SFF) policy. Then we built the $k$-d tree
of each $T_{\text{target}}^B$. For each possible example bundle
$b_{\text{example}}^A$, we first computed its dissimilarity
representation with the prototypes of $T_{\text{target}}^B$, then
segmented the target bundle $\hat{b}_{\text{target}}^B$ by querying
the $k$-d tree.

Following the common practice, see~\cite{garyfallidis2015robust}, as
accuracy of the estimation, we measured the degree of overlap between
$\hat{b}_{\text{target}}^B$ and the true target bundle
$b_{\text{target}}^B$, through the dice similarity coefficient (DSC)
at the voxel-level:
  $DSC = 2 \frac{|v(\hat{b}_{\text{target}}^B) \cap
    v(b_{\text{target}}^B)|}{|v(\hat{b}_{\text{target}}^B)| +
    |v(b_{\text{target}}^B)|} $
where $v(b)$ is the set of voxels crossed by the streamlines of bundle
$b$ and $|v(b)|$ is the number of voxels of $v(b)$.

The experiments were developed in Python code, on top of
DiPy\footnote{\url{http://nipy.org/dipy},
  \cite{garyfallidis2014dipy}.}. The code of all experiments is
available under a Free/OpenSource license
at~\url{http://github.com/emanuele/prni2017_comparison_of_distances}.

\subsection{Results}
In Table~\ref{tab:main_results}, we report the degree of overlap, as
mean DSC, obtained with the nearest neighbor supervised segmentation,
across the different tracts and the 8 different distance functions
considered. The mean is computed over all 90 pairs
$(b_{\text{example}}^A, T_{\text{target}}^B)$, obtained from the 10
subjects. For each bundle and distance function, we observed a
standard deviation of DSC of approximately $0.10$~\footnote{Which
  correspond to a standard deviation of the mean of $0.01$.}. Such
value includes the variances due to: the anatomical variability across
subjects, the limitations of the WMQL segmentation used as ground
truth and, in minor part, the approximation introduced by the
dissimilarity representation\footnote{Via bootstrap, we estimated an
  average contribution of $0.015$ to the value of the standard
  deviation of DSC.}.
  
\begin{table}
\centering
\caption{Mean DSC voxel table}
\label{tab:main_results}
\begin{tabular}{p{1.0cm} | p{0.5cm} | p{0.5cm} | p{0.5cm} | p{0.5cm} | p{0.5cm} | p{0.5cm} | p{0.5cm} | p{0.5cm}}
 & $d_{\text{\tiny MC}}$ & $d_{\text{\tiny SC}}$ & $d_{\text{\tiny LC}}$ & $d_{\text{\tiny MDF,12}}$ & $d_{\text{\tiny MDF,20}}$ & $d_{\text{\tiny MDF,32}}$ & $d_{\text{\tiny PDM}}$ & $d_{\text{\tiny varifolds}}$ \\
\hline
\hline
	 cg.left & 0.61 & 0.60 & 0.59 & 0.59 & 0.59 & 0.59 & 0.59 & 0.56 \\
	 \hline
	 cg.right & 0.60 & 0.59 & 0.58 & 0.58 & 0.57 & 0.58 & 0.57 & 0.55 \\
	 \hline
	 ifof.left & 0.49 & 0.48 & 0.47 & 0.48 & 0.48 & 0.47 & 0.48 & 0.49 \\
	 \hline
	 ifof.right & 0.47 & 0.46 & 0.45 & 0.45 & 0.45 & 0.45 & 0.46 & 0.44 \\
	 \hline
	 uf.left & 0.52 & 0.54 & 0.55 & 0.52 & 0.52 & 0.53 & 0.57 & 0.60 \\
	 \hline
	 uf.right & 0.49 & 0.52 & 0.51 & 0.49 & 0.49 & 0.49 & 0.52 & 0.56 \\
	 \hline
	 cc\_7 & 0.58 & 0.56 & 0.61 & 0.64 & 0.63 & 0.63 & 0.59 & 0.67 \\
	 \hline
	 cc\_2 & 0.49 & 0.50 & 0.52 & 0.53 & 0.53 & 0.54 & 0.57 & 0.59 \\
	 \hline
	 af.left & 0.51 & 0.49 & 0.51 & 0.51 & 0.50 & 0.50 & 0.52 & 0.50 \\
	 \hline
	 \hline
	 \textbf{means} & \textbf{0.53} & \textbf{0.53} & \textbf{0.53} & \textbf{0.53} & \textbf{0.53} & \textbf{0.53} & \textbf{0.54} & \textbf{0.55} \\
	 \hline
\end{tabular}
\end{table}

In Table~\ref{tab:times}, we report the time required by a modern
desktop computer to compute 90000 streamline-streamline distances
using the 8 distance functions considered in this study. The
differences in time are due to both the different computational cost
of the formulas in Section~\ref{sec:methods} and their
implementation. $d_{\text{MC}}$, $d_{\text{SC}}$, $d_{\text{LC}}$ and
$d_{\text{MDF}}$, available from DiPy, were implemented in
Cython. $d_{\text{PDM}}$ and $d_{\text{varifolds}}$ were implemented
by us in Python and NumPy\footnote{\url{http://www.numpy.org}}.

\begin{table}[h]
  \caption{Computational time for 90000 pairs of
    streamlines.}
  \label{tab:times}
  \centering
  \begin{tabular}{ p{0.7cm}| p{0.5cm}| p{0.5cm}| p{0.5cm}| p{0.5cm}| p{0.5cm}| p{0.5cm}| p{0.5cm}| p{0.5cm}}
 & $d_{\text{\tiny MC}}$ & $d_{\text{\tiny SC}}$ & $d_{\text{\tiny LC}}$ & $d_{\text{\tiny MDF,12}}$ & $d_{\text{\tiny MDF,20}}$ & $d_{\text{\tiny MDF,32}}$ & $d_{\text{\tiny PDM}}$ & $d_{\text{\tiny varifolds}}$ \\
    \hline
time(s) & $~0.5$ & $~0.5$  & $~0.5$  & $~0.03$  & $~0.04$  & $~0.05$  & $~16$  & $~28$ \\
  \hline
  \end{tabular} \\
\end{table}

\begin{figure*} [!ht]
\centering
\begin{minipage} [b] {0.29\textwidth}
\centering
\centering
  \includegraphics[width=3.765cm]{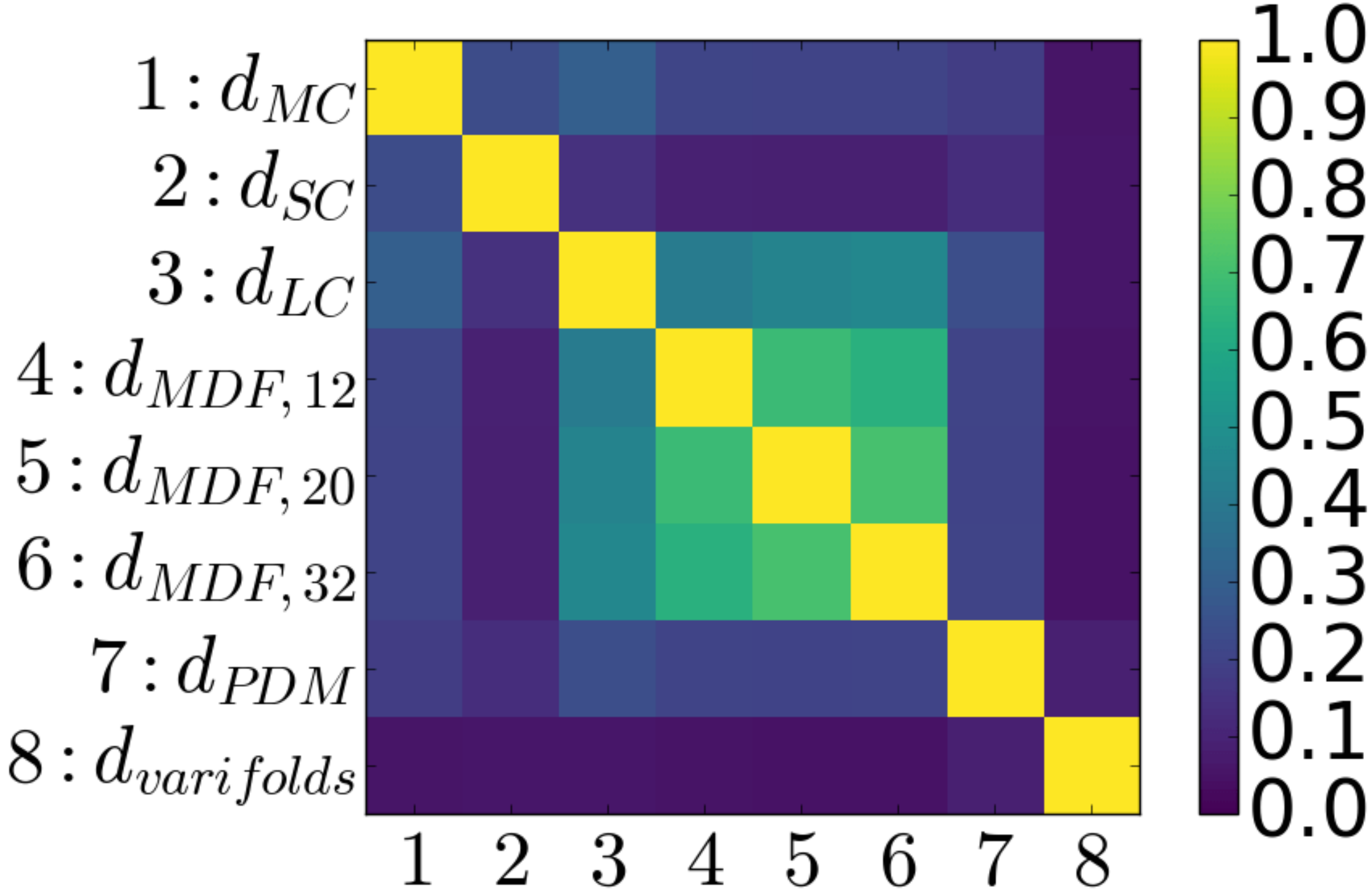} 
  \caption{Frequency with which two distance functions selected the
    same nearest neighbor of a streamline during all our experiments.}
  \label{fig:cross_overlap_ss}
\end{minipage}
\hspace{7mm}
\begin{minipage} [b] {0.65\textwidth}
\centering
\subfigure 
{\includegraphics[width=0.235\columnwidth]{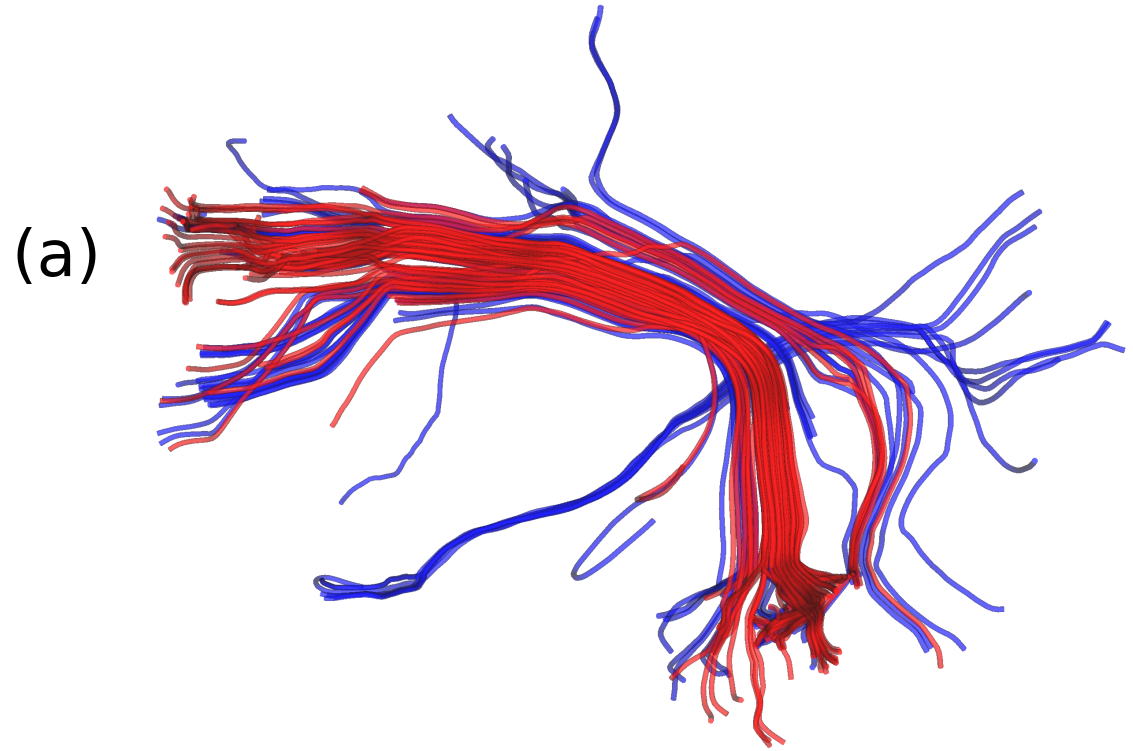}}
\hspace{0mm}
\subfigure 
{\includegraphics[width=0.235\columnwidth]{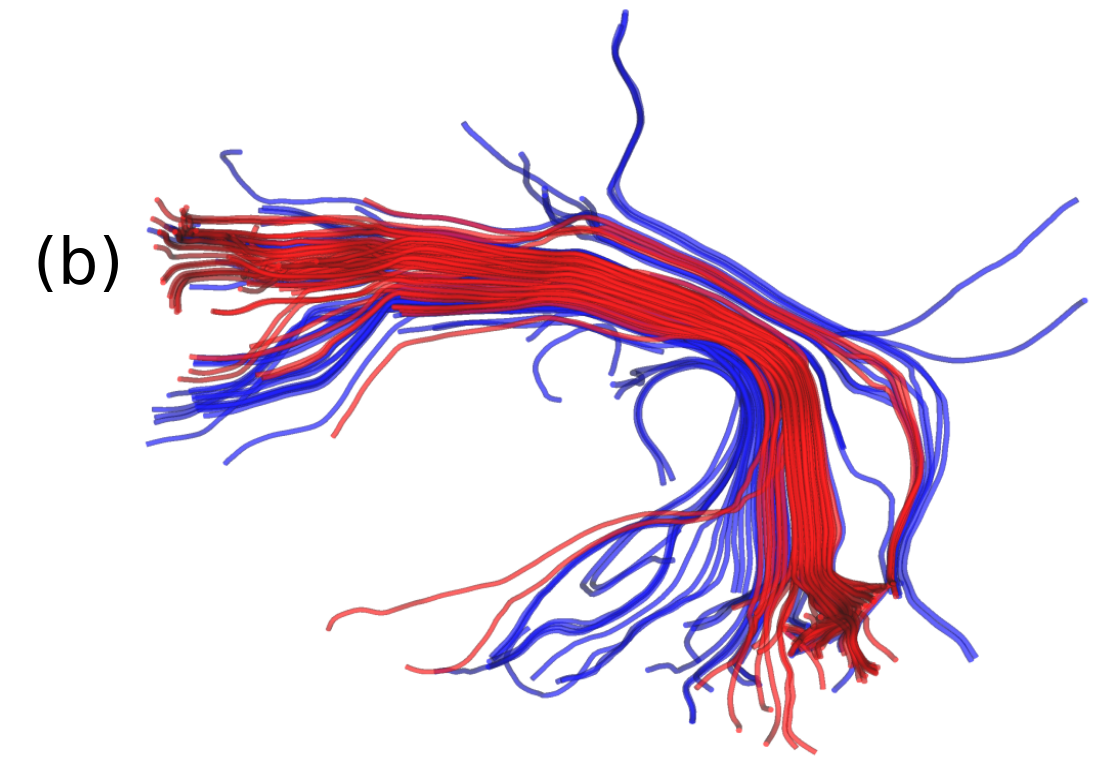}}
\hspace{0mm}
\subfigure 
{\includegraphics[width=0.235\columnwidth]{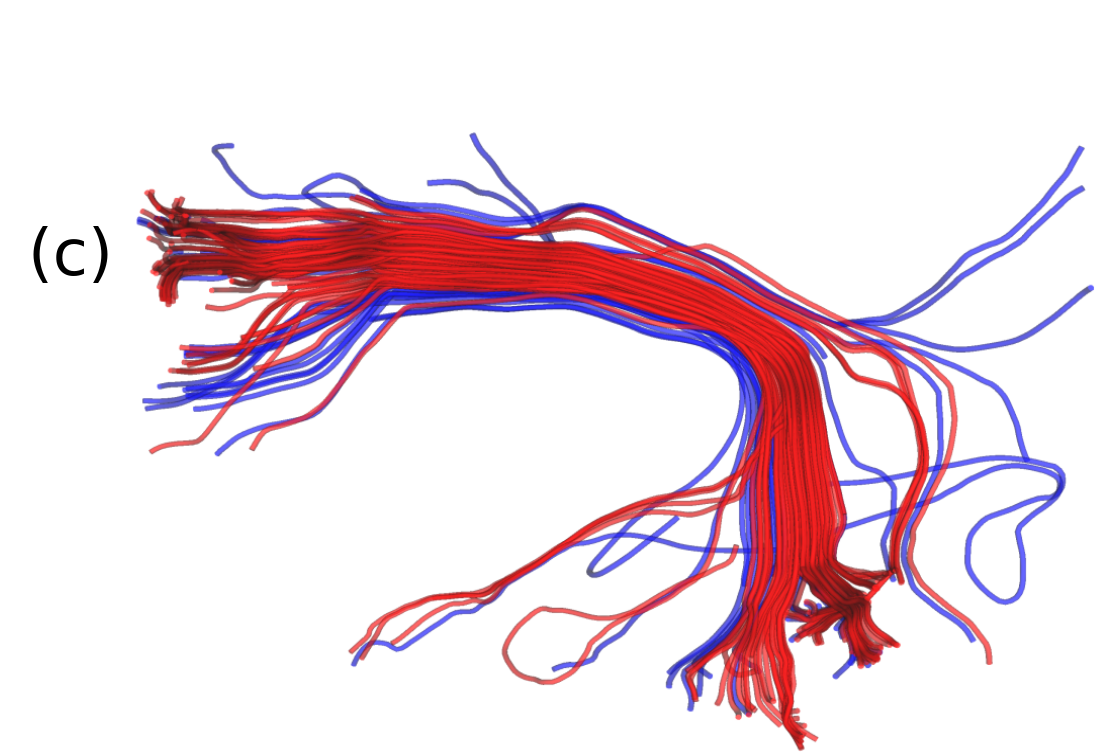}}
\hspace{0mm}
\subfigure 
{\includegraphics[width=0.235\columnwidth]{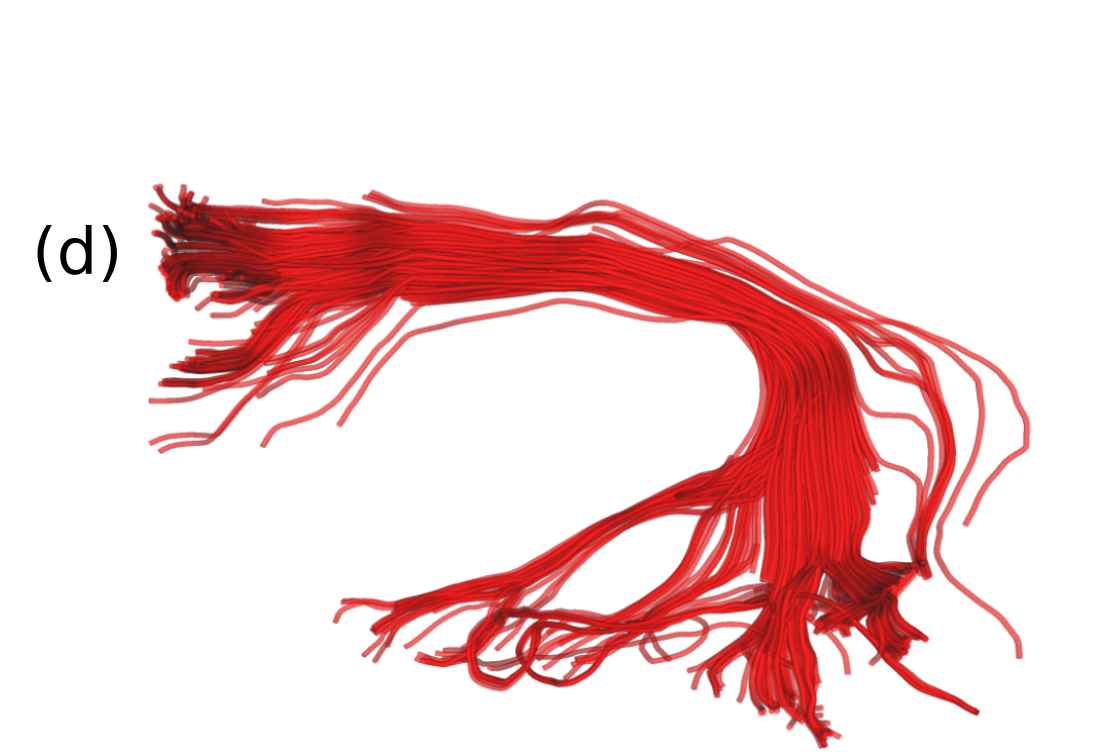}}
\caption{Example of segmented arcuate fasciculus left with NN using (a) $d_{\text{\tiny LC}}$ (DSC=0.64) (b) $d_{\text{\tiny MDF,20}}$ (DSC=0.69) and (c) $d_{\text{\tiny varifolds}}$ (DSC=0.71). (d) Ground truth arcuate fasciculus left. True positive streamlines in red and false positives in blue. Subject A: HCP ID 201111, subject B: HCP ID 124422.}
\label{fig:af_left1}
\end{minipage}
\end{figure*}

In order to collect more insight on the results of
Table~\ref{tab:main_results}, we investigated in
Figure~\ref{fig:cross_overlap_ss} whether different distance functions
returned the same nearest neighbor streamlines. We expect that
distance functions, that are based on different geometric principles,
have a different nearest neighbor. In
Figure~\ref{fig:cross_overlap_ss}, each entry represents the frequency
with which two distance functions returned the same nearest neighbor
of a given streamline. Such frequency is computed over all streamlines
of all tracts of all pairs of subjects considered in the experiments,
i.e. approximately 200000 nearest neighbor computations.

%
%

%% file: discussion.tex
\section{Discussion and Conclusion}
\label{sec:discussion}
The results reported in Table~\ref{tab:main_results} clearly show that
there are no major differences in the accuracy of the supervised
segmented bundles, measured as DSC, when using different distance
functions. The highest mean DSC value, i.e. $0.55$ for
$d_{\text{varifolds}}$, is not significantly higher than the other
values. This is partly different from the results reported
in~\cite{siless2013comparison} but, as mentioned in
Section~\ref{sec:introduction}, that work investigated segmentation as
\emph{unsupervised} clustering of streamlines, while we focus on
\emph{supervised} bundle segmentation. The supervised approach is
example-based, thus directly driven by anatomy, while clustering is
not. For this reason, differences in the results of the two approaches
are to be expected.

The results in Figure~\ref{fig:cross_overlap_ss} show that different
distance functions often result in different nearest neighbor of a
streamline, with some exceptions. Expectedly, all MDF distance
functions frequently select the same nearest neighbor, $\approx$65\%
of the times. Surprisingly, $d_{\text{LC}}$ agrees with them
$\approx$45\% of the times. In all other cases the agreement is very
low, between $5\%$ and $25\%$.

Why do different nearest neighbors lead to a similar quality
of segmentation? The potential disagreement between the results in
Table~\ref{tab:main_results} and Figure~\ref{fig:cross_overlap_ss} can
be explained by the following argument. At the \emph{local} level,
different distances clearly have a geometrically different concept of
proximity, frequently leading to different nearest
neighbors. Nevertheless, we observed that such different neighbors do
not lie far apart from each other so, at a \emph{higher/aggregated}
level of bundle, it should not be a surprise that they lead to a
comparable quality of segmentation. 
This can also be seen in Figure~\ref{fig:af_left1}, where the false positives of the bundles segmented with different distances are almost the same, while the false negatives are different.
Moreover, Table~\ref{tab:main_results} presents a \emph{voxel} measure of bundle
overlap, while Figure~\ref{fig:cross_overlap_ss} presents a
\emph{streamline} measure. A voxel-based measure of bundle overlap is
inherently less sensitive than a streamline-based measure, because
different proximal streamlines usually have many voxels in common. So
when two distance functions lead to different (but proximal) nearest
neighbors, they will positively contribute in terms of voxel overlap,
but not in terms of streamline overlap.



Furthermore, we observe in Table~\ref{tab:times} that the
computational times of the distance functions can be very
different. For instance, there are more than two orders of magnitude
between the computational time of $d_{\text{MDF}}$ and the one of
$d_{\text{varifolds}}$. To conclude, for the supervised segmentation
task based on a voxel-based measure, we suggest that practitioners
prefer fast distance functions, such as $d_{\text{MDF}}$,
$d_{\text{MC}}$, $d_{\text{SC}}$ or $d_{\text{LC}}$, over slower ones,
like $d_{\text{PDM}}$ and $d_{\text{varifolds}}$.



%% file: olivetti_streamline_distance.bbl
\begin{thebibliography}{10}
\providecommand{\url}[1]{#1}
\csname url@samestyle\endcsname
\providecommand{\newblock}{\relax}
\providecommand{\bibinfo}[2]{#2}
\providecommand{\BIBentrySTDinterwordspacing}{\spaceskip=0pt\relax}
\providecommand{\BIBentryALTinterwordstretchfactor}{4}
\providecommand{\BIBentryALTinterwordspacing}{\spaceskip=\fontdimen2\font plus
\BIBentryALTinterwordstretchfactor\fontdimen3\font minus
  \fontdimen4\font\relax}
\providecommand{\BIBforeignlanguage}[2]{{%
\expandafter\ifx\csname l@#1\endcsname\relax
\typeout{** WARNING: IEEEtran.bst: No hyphenation pattern has been}%
\typeout{** loaded for the language `#1'. Using the pattern for}%
\typeout{** the default language instead.}%
\else
\language=\csname l@#1\endcsname
\fi
#2}}
\providecommand{\BIBdecl}{\relax}
\BIBdecl

\bibitem{catani2015atlas}
M.~Catani and M.~T. de~Schotten, \emph{{Atlas of Human Brain Connections}},
  1st~ed.\hskip 1em plus 0.5em minus 0.4em\relax Oxford University Press, Apr.
  2015.

\bibitem{catani2002virtual}
M.~Catani, R.~J. Howard, S.~Pajevic, and D.~K. Jones, ``{Virtual in vivo
  interactive dissection of white matter fasciculi in the human brain.}''
  \emph{NeuroImage}, vol.~17, no.~1, pp. 77--94, Sep. 2002.

\bibitem{garyfallidis2012quickbundles}
E.~Garyfallidis, M.~Brett, M.~M. Correia, G.~B. Williams, and I.~Nimmo-Smith,
  ``{QuickBundles, a Method for Tractography Simplification.}'' \emph{Frontiers
  in neuroscience}, vol.~6, 2012.

\bibitem{yoo2015example}
S.~W. Yoo, P.~Guevara, Y.~Jeong, K.~Yoo, J.~S. Shin, J.-F. Mangin, and J.-K.
  Seong, ``{An Example-Based Multi-Atlas Approach to Automatic Labeling of
  White Matter Tracts},'' \emph{PloS one}, vol.~10, no.~7, 2015.

\bibitem{sharmin2016alignment}
N.~Sharmin, E.~Olivetti, and P.~Avesani, ``{Alignment of Tractograms as Linear
  Assignment Problem},'' in \emph{Computational Diffusion MRI}.\hskip 1em plus
  0.5em minus 0.4em\relax Springer, 2016, pp. 109--120.

\bibitem{corouge2006fiber}
I.~Corouge, P.~Fletcher, S.~Joshi, S.~Gouttard, and G.~Gerig, ``{Fiber
  tract-oriented statistics for quantitative diffusion tensor MRI analysis},''
  \emph{Medical Image Analysis}, vol.~10, no.~5, pp. 786--798, Oct. 2006.

\bibitem{batchelor2006quantification}
P.~G. Batchelor, F.~Calamante, J.~D. Tournier, D.~Atkinson, D.~L.~G. Hill, and
  A.~Connelly, ``{Quantification of the shape of fiber tracts},''
  \emph{Magnetic Resonance in Medicine}, vol.~55, no.~4, pp. 894--903, Apr.
  2006.

\bibitem{garyfallidis2015robust}
E.~Garyfallidis, O.~Ocegueda, D.~Wassermann, and M.~Descoteaux, ``{Robust and
  efficient linear registration of white-matter fascicles in the space of
  streamlines},'' \emph{NeuroImage}, vol. 117, pp. 124--140, Aug. 2015.

\bibitem{gori2016parsimonious}
P.~Gori, O.~Colliot, L.~Marrakchi-Kacem, Y.~Worbe, F.~De~Vico~Fallani,
  M.~Chavez, C.~Poupon, A.~Hartmann, N.~Ayache, and S.~Durrleman,
  ``{Parsimonious Approximation of Streamline Trajectories in White Matter
  Fiber Bundles.}'' \emph{IEEE transactions on medical imaging}, Jul. 2016.

\bibitem{moberts2005evaluation}
B.~Moberts, A.~Vilanova, and J.~J. van Wijk, ``{Evaluation of Fiber Clustering
  Methods for Diffusion Tensor Imaging},'' in \emph{VIS 05. IEEE Visualization,
  2005.}\hskip 1em plus 0.5em minus 0.4em\relax IEEE, 2005, pp. 65--72.

\bibitem{siless2013comparison}
V.~Siless, S.~Medina, G.~Varoquaux, and B.~Thirion, ``{A Comparison of Metrics
  and Algorithms for Fiber Clustering},'' in \emph{2013 International Workshop
  on Pattern Recognition in Neuroimaging}.\hskip 1em plus 0.5em minus
  0.4em\relax IEEE, Jun. 2013, pp. 190--193.

\bibitem{olivetti2016alignment}
E.~Olivetti, N.~Sharmin, and P.~Avesani, ``{Alignment of Tractograms As Graph
  Matching},'' \emph{Frontiers in Neuroscience}, vol.~10, 2016.

\bibitem{sotiropoulos2013effects}
S.~N. Sotiropoulos, S.~Moeller, S.~Jbabdi, J.~Xu, J.~L. Andersson, E.~J.
  Auerbach, E.~Yacoub, D.~Feinberg, K.~Setsompop, L.~L. Wald, and Others,
  ``{Effects of image reconstruction on fiber orientation mapping from
  multichannel diffusion MRI: reducing the noise floor using SENSE},''
  \emph{Magnetic resonance in medicine}, vol.~70, no.~6, pp. 1682--1689, 2013.

\bibitem{wassermann2013describing}
D.~Wassermann, N.~Makris, Y.~Rathi, M.~Shenton, R.~Kikinis, M.~Kubicki, and
  C.-F.~F. Westin, ``{On describing human white matter anatomy: the white
  matter query language.}'' \emph{Medical image computing and computer-assisted
  intervention : MICCAI ... International Conference on Medical Image Computing
  and Computer-Assisted Intervention}, vol.~16, no. Pt 1, pp. 647--654, 2013.

\bibitem{zhang2008identifying}
S.~Zhang, S.~Correia, and D.~H. Laidlaw, ``{Identifying White-Matter Fiber
  Bundles in DTI Data Using an Automated Proximity-Based Fiber-Clustering
  Method},'' \emph{IEEE Transactions on Visualization and Computer Graphics},
  vol.~14, no.~5, pp. 1044--1053, Sep. 2008.

\bibitem{charon2013varifold}
N.~Charon and A.~Trouv\'{e}, ``{The Varifold Representation of Nonoriented
  Shapes for Diffeomorphic Registration},'' \emph{SIAM Journal on Imaging
  Sciences}, vol.~6, no.~4, pp. 2547--2580, Jan. 2013.

\bibitem{olivetti2012approximation}
E.~Olivetti, T.~B. Nguyen, and E.~Garyfallidis, ``{The Approximation of the
  Dissimilarity Projection},'' \emph{IEEE Intl Workshop on Pattern Recognition
  in NeuroImaging}, vol.~0, pp. 85--88, 2012.

\bibitem{pekalska2005dissimilarity}
E.~Pekalska and R.~P.~W. Duin, \emph{{The Dissimilarity Representation for
  Pattern Recognition: Foundations And Applications (Machine Perception and
  Artificial Intelligence)}}.\hskip 1em plus 0.5em minus 0.4em\relax World
  Scientific Publishing Company, Dec. 2005.

\bibitem{bentley1975multidimensional}
J.~L. Bentley, ``{Multidimensional binary search trees used for associative
  searching},'' \emph{Communications of the ACM}, vol.~18, no.~9, pp. 509--517,
  1975.

\bibitem{vanessen2013wuminn}
D.~C. Van~Essen, S.~M. Smith, D.~M. Barch, T.~E.~J. Behrens, E.~Yacoub, and
  K.~Ugurbil, ``{The WU-Minn Human Connectome Project: An overview},''
  \emph{NeuroImage}, vol.~80, pp. 62--79, Oct. 2013.

\bibitem{tournier2007robust}
J.-D. Tournier, F.~Calamante, and A.~Connelly, ``{Robust determination of the
  fibre orientation distribution in diffusion MRI: Non-negativity constrained
  super-resolved spherical deconvolution},'' \emph{NeuroImage}, vol.~35, no.~4,
  pp. 1459--1472, May 2007.

\bibitem{garyfallidis2014dipy}
E.~Garyfallidis, M.~Brett, B.~Amirbekian, A.~Rokem, S.~van~der Walt,
  M.~Descoteaux, I.~Nimmo-Smith, and D.~Contributors, ``{Dipy, a library for
  the analysis of diffusion MRI data},'' \emph{Frontiers in Neuroinformatics},
  vol.~8, no.~8, pp.~1+, Feb. 2014.

\end{thebibliography}
